\documentclass[10pt]{article}
\usepackage[letterpaper]{geometry}
\usepackage{hicss}
\usepackage{times}
\usepackage[none]{hyphenat}
\usepackage{url}
\usepackage{latexsym}
\usepackage{listings}
\usepackage{indentfirst}
\usepackage{graphicx}
\graphicspath{{images/}}

\usepackage{hyperref}
\usepackage{fancyhdr,graphicx,amsmath,amssymb}
\usepackage{array}
\usepackage[ruled,vlined]{algorithm2e}
\usepackage{multirow}
\usepackage[utf8]{inputenc}
\usepackage[svgnames, table]{xcolor}
\usepackage{array, booktabs, makecell, multirow}
\usepackage{enumitem}
\usepackage{listings}
\newcolumntype{P}[1]{>{\centering\arraybackslash}p{#1}}
\emergencystretch=\maxdimen
\title{Unified Explanations in Machine Learning Models: A Perturbation Approach}

\author{Jacob Dineen \\
 Arizona State University \\
 {\underline{jdineen@asu.edu}} \\\And
 Don Kridel \\
 U Missouri, St Louis \\
 {\underline{dkridel@gmail.com} }\\\And 
 Daniel Dolk \\
 Naval Postgrad School \\
 {\underline{drdolk@nps.edu}} \\\And 
  David Castillo \\
 JP Morgan Chase \\
 {\underline{dcastilloaz@gmail.com}} \\
 }

\date{}

\begin{document}
\maketitle

\begin{abstract}
A high-velocity paradigm shift towards Explainable Artificial Intelligence (XAI) has emerged in recent years. Highly complex Machine Learning (ML) models have flourished in many tasks of intelligence, and the questions have started to shift away from traditional metrics of validity towards something deeper: What is this model telling me about my data, and how is it arriving at these conclusions? Inconsistencies between XAI and modeling techniques can have the undesirable effect of casting doubt upon the efficacy of these explainability approaches. To address these problems, we propose a systematic, perturbation-based analysis against a popular, model-agnostic method in XAI, SHapley Additive exPlanations (Shap). We devise algorithms to generate relative feature importance in settings of dynamic inference amongst a suite of popular machine learning and deep learning methods, and metrics that allow us to quantify how well explanations generated under the static case hold. We propose a taxonomy for feature importance methodology, measure alignment, and observe quantifiable similarity amongst explanation models across several datasets. 


\end{abstract}

\section{Introduction}

A long-standing problem in predictive analytics has been the disconnect between modelers (statisticians, mathematicians, and data scientists) at the model development stage and end-users at the organizational and decision-maker levels. The latter group is whom the models are ostensibly built for in the first place, but typically does not have the analytical background necessary for a full comprehension of the resulting model artifacts whereas the former community may lack the requisite knowledge of the problem domain driving the requirements of decision-makers. This “cultural divide” too frequently has the undesirable consequence of diminishing the utility of modeling to its intended audience. 

This communication problem has migrated into the arena of machine learning (ML) and artificial intelligence (AI) in recent times, giving rise to the need for and subsequent emergence of Explainable AI (XAI). XAI has arisen from growing discontent with “black box” models, often in the form of neural networks and other emergent, dynamic models (e.g., agent-based simulation, genetic algorithms) that generate outcomes lacking in transparency. This has also been studied through the lens of general machine learning, where classic methods also face an interpretability crisis for high dimensional inputs \cite{lipton2018mythos}. Applications such as facial recognition have been met with stern resistance as, too often, mistaken identifications have led to unnecessary and serious disruption in individuals’ lives \cite{garvie2016facial,miller2020matter,georgopoulos2020investigating}. As another example, the pre-existing bias in historic data-sets used for building ML algorithms (MLAs) has resulted in some people or marginalized communities having firsthand experience with algorithmic unfairness \cite{johnson2019artificial, corbett2018measure}. This is a very sensitive issue for companies for whom the public perception of policy fairness and impartiality is critical to their business and well-being, and has paved the way for increased research interest in algorithmic fairness and equality. 

A recent well-received book by John Kay and Mervyn King, “Radical Uncertainty: Decision-Making Beyond the Numbers” \cite{radical} highlight similar problems for economic models. The authors suggest a need for reference narratives, which are stories that can be marshaled to address the overriding objective of unraveling “what’s going on here?” We adopt this perspective as our long-term strategy for model explainability. 

Model transparency is and will continue to be a growing area of interest as ML/AI models continue to develop, and organizations and users will continue to demand improved accountability. Translating mathematical and data science expertise into decision-making expertise remains a significant obstacle in gaining organizational acceptance of model artifacts. We believe that advances in model explainability and interpretation are essential to bridge this gap.\\

We make the following contributions here:
\begin{enumerate}
    \item \textbf{Algorithm for Dynamic Feature Perturbation:} We introduce algorithm \ref{algo : 1} for dynamic perturbation of a testing set akin to evasion attacks in adversarial ML literature. The algorithm performs, iteratively, continuous and categorical feature perturbation and measures sensitivity on the model's output.
    \item \textbf{Metric Derivation:} We formulate and adapt two distance-based metrics to systematically quantify relative feature importance under our proposed algorithms.
    \item \textbf{Evaluation and Comparison:} We compare the similarities between a well-known XAI technique in Shap \cite{Lundberg_Lee_2017} to our proposed method and analyze harmony and/or disharmony between the static and dynamic case.
\end{enumerate}

\section{Background}
\subsection{Preliminaries}
While some work in XAI focuses on predictive model explainability on the static part of the process, we focus on contributions to both the static and dynamic scenarios. We define these terms under the following taxonomy:
\begin{itemize}
  \item \emph{Static Scenarios:} Given static, partitioned training and testing sets $\in X, Y$, identify feature importances (FIs) using Shap, LIME, relative entropy, model weights, log odds, etc. Under the static case, we generate FIs that allow us to understand the decision boundaries being drawn under the model fitting process.
  \item \emph{Dynamic Scenarios:} Under prediction scenarios, the effect, or sensitivity, of the model's generalizability when instances of the testing set are artificially perturbed $X^{test'}$ such that they are likely to be out-of-sample against the data the model was fitted against, e.g., $f(\theta, X^{test'})$. 
\end{itemize}

There is a likeness to alternative nomenclature used in the field of Adversarial Machine Learning (AML), where a similar taxonomy is proposed by \cite{Xu2020AdversarialAA}. Our framework could be seen as being loosely analogous with evasion attacks under the AML setting, under the guise of sensitivity analysis. \emph{Evasion attacks} are adversarial attacks where the underlying attack occurs after the specified learning algorithm is fully trained, e.g., the architecture and the learnable parameters are fixed and immutable.

\subsection{Related Work}
Previous work \cite{dolk2020model} explored the taxonomy noted above to analyze how well measures of feature importance hold up under 'what-if' perturbations. We extend this work, not in breadth, but depth, offering a framework to systematically quantify the similarity between the two. 

Within the literature on XAI, densely populated in recent years, are methods derived explicitly to counter the black-box nature of Deep Neural Networks \cite{sundararajan2017axiomatic,shrikumar2017learning,binder2016layer}, particularly Convolutional Neural Networks (CNNs) used for computer vision applications. \cite{Lundberg_Lee_2017, ribeiro2016should} are seen as more generalized means of attributing explanations to model predictions, but LIME focuses on the case of local, linear approximations and SHapley Additive exPlanations (Shap) \footnote{https://github.com/slundberg/shap}offers a more comprehensive effort towards exploring local and global model explanations. Our focus in this paper is on Shap, which follows from its longstanding impact in literature and applications, the attractiveness of the properties of additive feature attributions, as well as the user studies that noted consistency between model explanations and human explanations. 

\emph{Shapley Regression Values (SRGs)}, EQ \ref{eq:shapreg}, generate FIs in the presence of multicollinearity. Generating SRGs is an iterative, expensive process that involves training a model on every possible combination of feature subsets and measuring the overall effect of the model of occluding a specific feature \cite{Lipovetsky2001AnalysisOR}. $\phi$ is the output SRGs as a weighted average of differences between all such possible subsets of features $S \subseteq F$.

\begin{equation}
\begin{array}{l}
\phi_{i}=\sum_{S \subseteq F \backslash\{i\}} \frac{|S| !(|F|-|S|-1) !}{|F| !} \cdot \\\quad \quad \quad \quad \quad \quad \quad \quad \left[f_{S \cup\{i\}}\left(x_{S \cup\{i\}}\right)-f_{S}\left(x_{S}\right)\right]
\label{eq:shapreg}
\end{array}
\end{equation}

\begin{figure}[h!]
\centering
\includegraphics[scale=0.65]{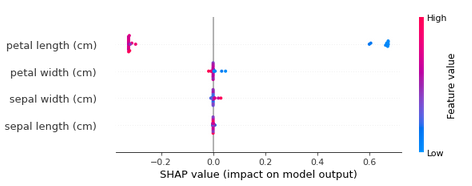}
\caption{Shap can be used to generate local or global explanations about each label $y_i \in Y$. The global explanation for the $0^{th}$ class of the Iris Flower Dataset is shown above. The graph shows us that when looking at predictions, globally, made for the $0th$ class, our model is generally not using petal width, sepal width, or sepal length to draw a decision boundary. Petal length SRGs are more dispersed and are generally showing that a high value is more indicative of a lower $p(y_i = 0)$, and a lower value has a higher influence on mapping it to the $0^{th}$ class $p(y_i = 1)$.}
\label{fig:shap}
\end{figure}

A turn from its original use case in cooperative game theory, where marginal contributions, or losses, are distributed between coalitions, there is a naturalness to applying it to predictive models, where input feature subsets are considered as coalitions and the Shapley value is the contribution (gain, loss) against the explanatory power of the model \cite{iooss2019shapley}. We show an example of Shap usage for global explanations in Figure \ref{fig:shap}, where we adopt the term 'feature importance graph' for end users.  We partly rely on this formulation for dynamic perturbation, but do not emphasize the presence, or lack thereof, of features, instead focusing on their magnitudinal effect given a previously trained model $f$.

\section{Methodology}
We note our algorithm for dynamic perturbation in algorithm \ref{algo : 1} in the Appendix. All code for the algorithm and experiments is open sourced \href{https://github.com/jacobdineen/hiccs2021}{here}. We let $X \in R^{nxm}$ denote a matrix where $X_{i,j}$ represents the $jth$ feature value of the $ith$ datapoint in $X$. Let the cardinality of the column dimension be represented by $|m|$, e.g., let $X(j \in |m|)$ represent the $jth$ column of $X$. $p_i \in P$ is a perturbation parameter.

\emph{The Continuous Case:} Given a trained model $f$, a continuous feature $j$, and a perturbation parameter $p_i$, apply the perturbation parameter to $X^{test}$: $X^{test'} = X^{test}(j) * p_i$, to arrive at a perturbed testing set. Perform inference on $X^{test'}$ using the original $f$ to generate predictions: $\hat y = f(X^{test'})$. Apply \textit{any} metric $m$ to measure predictions against actuals $m(\hat y, y)$

\emph{The Categorical Case:} The Categorical case has the same underlying mechanics, but $p_i$ is no longer a multiplicative scaling factor. In this case $p_i$ is a factor influencing the presence of a categorical variable. $|X(j)| = n$, and $|X(j = 1)|$ is the length of the set where the categorical feature is active. Applying the perturbation parameter $p_i$ to $X^{test}$: $X^{test'} = X^{test}(j) * p_i$, we scale the presence of data observations with an activated (boolean) categorical feature. When $p_i = 2$ we have doubled the size of $|X(j = 1)|$. We make the assumption that all categorical features are boolean, as ordinal variables can be transformed into one-hot representations with binary responses.

\subsection{Absolute Normalized Shap}
To compute the Relative Feature Importance (RFIs) for each feature $j \in |m|$ given Shap values, we propose measuring each feature's absolute contribution to the absolute total contribution of all Shap values for a datapoint's correct class. Formally, we represent this as:

\begin{equation}
S(j) = \frac{\mid \text { shap }\left(X(j)\right) \mid}{\sum_{j}^{m}\left|\operatorname{shap}\left(X(j)\right)\right|}
\label{eq: Absolute Normalized Shap}
\end{equation}

\subsection{Absolute Normalized Weighted Average (ANWA)}

Let EQ \ref{eq: ANWA} represent the weighted average given a feature $x_i \in X$ and a scaling factor (perturbation parameter) $p_i \in P$, where $P \in (0,2)$ to negate the effect of vanishing weights. Let $f(X^{test'}, P)$ be the ensuing model output from applying the scaling factor $p_i$ to the feature column $X(j)$.

\begin{equation}
W\left(j, P\right)=\frac{\sum_{i}^{|P|}w_i \cdot f\left(X(j), p_{i}\right)}{\sum_{i}^{|P|} w_i}
\label{eq: ANWA}
\end{equation}

The term under the summation in the numerator $w_i=(1-\left|1-p_{i}\right|)$ places more weight on perturbations closer to the unperturbed base case, i.e., when $p_i = 0$, we have our base case $f(X^{test})$. We present a working example of this in Table \ref{table: tab1}.

\begin{table}[h!]
\small
    \centering
    \setlength\tabcolsep{14pt}
    \begin{tabular}{|l|l|l|l|}
    \hline
        \textbf{$p_i$} & \textbf{$w_(p_i)$} & \textbf{$F(X')$} & Contribution \\ \hline
        0.01 & 0.01 & 0.5 & 0.005 \\ \hline
        0.5 & 0.5 & 0.83 & 0.42 \\ \hline
        1 & 1 & 0.96 & 0.96 \\ \hline
        1.5 & 0.5 & 0.7 & 0.35 \\ \hline
        1.99 & 0.01 & 0.36 & 0.004 \\ \hline
         &  &  \textbf{ANWA} & \textbf{0.863} \\ \hline
    \end{tabular}
    \caption{The table can be read as follows: Column $p_i$ represents the perturbation parameter, $w(p_i)$ is the weight we place on the predictions on the dataset perturbed by $p_i$, $F(X')$ is the models output (given a metric) on the perturbed dataset, and the ANWA is in the lower right cell as a linear combination of the previous two columns. More weight is placed on predictions where $X'$ is closer to the original, unperturbed dataset $X$.}
    \label{table: tab1}
\end{table}

Let $u(\cdot) = |f(X^{test}) - f(X^{test'}, P)|$ represent the absolute difference between the base case and the weighted average $W$ induced from perturbations $p_i \in P$ on $x_i$. We compute the relative importance of feature $x_i$ on the model's output (accuracy/precision/recall/f1) as

\begin{equation}
I\left(x_{i}, P\right)=\frac{u\left(x_{i}, P \right)}{\sum_{i}^{n} u\left(x_{i}, P\right)}
\label{eq: 1}
\end{equation}

This outputs the contribution of a model's performance on a perturbed dataset given $p_i$ against the total contribution generated via $\forall p \in P$, or the absolute difference between the base case and the weighted average over perturbing a feature $|P|$ times, over the sum of absolute differences from repeating this for $X(j), j \in |m|$. 

\subsection{Metric Comparison}
To quantify similarity between RFI vectors associated with the outputs from EQs \ref{eq: Absolute Normalized Shap} and \ref{eq: ANWA}, we leverage two popular distance metrics used in literature for a variety of tasks. The proposed usage of varying distance metrics helps to shape the underlying reference narrative of the model artifacts, where some end users may want to drill down on the full explanation model, and others may side with \textit{exclusively} focusing on the most important features.


\textbf{\emph{Cosine Similarity}}
Cosine similarity is a distance metric bounded between $[0,1]$, where a cosine value of 0 means that two vectors are orthogonal to one another and have no intrinsic similarity, while a cosine value approaching 1 indicates a greater likeness between vectors. In our case, a higher cosine value represents a greater harmony between explanations under the proposed taxonomy of static and dynamic explainability.

\begin{equation}
\operatorname{dist}(A, B)=\cos (\theta)=\frac{A \cdot B}{\|A\| \|B \|}
\label{eq: cos}
\end{equation}

Let $\Vec{i}$ be a vector corresponding to the RFIs computed using EQ \ref{eq: 1}, and let $\Vec{j}$ be a vector containing the proportion of summed absolute Shap values against the whole population (all features).

\begin{equation}
\begin{array}{l}
    \Vec{i}=<I\left(x_{i}, X, P\right), \ldots, I\left(x_{n}, X, P\right)> \\
    \Vec{j}=<S(x_i), ..., S(x_n)>
\end{array}
\end{equation}

We superpose these vectors into EQ \ref{eq: cos} to arrive at the cosine similarity between the RFIs generated under the static and dynamic cases.

\begin{equation}
\operatorname{cos}(\Vec{i}, \Vec{j})=\frac{\Vec{i} \cdot \Vec{j}}{\|\Vec{i}\| \mid \Vec{j} \|}
\end{equation}

\textbf{\emph{Jaccard Similarity}}  Jaccard Similarity (JS) is a popular metric when looking at top-k performance commonly used to measure efficacy in Recommender System \cite{Al-Shamri_2014, ayub2018jaccard} or Multi-label classification \cite{gouk2016learning, montanes2014dependent} tasks. We rationalize our inspection of similarity under JS with the assumption that the magnitude of the RFIs is not as important as their ranking in some cases, which is often the case when generating reason codes in regulated financial applications. We ask the question: "How often do FI rankings align under static and dynamic scenarios?". Like Cosine Similarity, this metric is also bounded, i.e., $0 \leq J(A, B) \leq 1$, making it an attractive option for analysis. 

\begin{equation}
J(A, B)=\frac{|A \cap B|}{|A \cup B|}=\frac{|A \cap B|}{|A|+|B|-|A \cap B|}
\label{metric: jacc}
\end{equation}

In EQ \ref{metric: jacc}, we see that $J$ measures the cardinality of the intersection of two sets against the union of the two sets. Here, we consider the two sets, $A(k)$ and $B(k)$ to be the top-k ranked RFIs given EQs \ref{eq: Absolute Normalized Shap} and \ref{eq: ANWA}. 

\emph{An Example:} Suppose we are looking at the Jaccard Similarity for the top-2 highest RFI features given an explanation model's output. Let $A$ be the set containing the top-2 most important features under EQ \ref{eq: Absolute Normalized Shap} and let $B$ be the top-2 ranked RFIs under EQ \ref{eq: ANWA} (refer to Figure \ref{fig:shapddd}). We compute JS here: 

\begin{equation}
\begin{array}{l}
    A(k = 2) = \{\text{petal length, petal width}\} \\
    B(k = 2) = \{\text{petal width, sepal length}\} \\
    J(A, B)=\frac{1}{2+2-1} = \frac{1}{3}
\end{array}
\end{equation}

\section{Experiments}
\begin{table*}
    \small
    \centering
    
    \makebox[\textwidth][c]{%

    \begin{tabular}{|l|l|l|l|l|l|l|l|}
    \hline
        \textbf{Dataset} & \textbf{Reference} & \textbf{Size} & \textbf{Ind. Features} & \textbf{Continuous}  & \textbf{Categorical} & \textbf{Classification} & \textbf{Classes} \\ \hline
        Iris & \cite{Dua:2019} & 150 & 4 & X &  & X & 3\\ \hline
        Wine & \cite{Dua:2019} & 177 & 13 & X &  & X & 10 \\ \hline
        Breast Cancer & \cite{Dua:2019} & 568 & 30 & X &  & X & 2 \\ \hline
        UCI Census & \cite{Dua:2019}  & 48842 & 14 & X & X & X & 2\\ \hline
        Synthetic Fraud & \cite{jpm1, jpm2}  & 3430 & 8 & X & X & X & 2\\ \hline

    \end{tabular}
    }
    \caption{Summary of the Datasets we explore. Each falls under the bucket of a classification task, where a data point consists of an $(x,y)$ tuple and where $y$ is a discrete variable. Classes is a value corresponding to the number of discrete values in the target variable. All datasets are split 80/20 into train and test sets with randomized shuffling. Datasets with categorical features are run through a label encoder to generate one hot representations. Further details on preprocessing are available in the Appendix.}
    \label{table : datasets}
\end{table*}

Our focus in this paper is to systematically analyze XAI under static and dynamic scenarios. Analysis of model performance is beyond the scope of this paper, and as such we utilize known classification datasets in the literature that \textit{most} existing methods can solve. We run our experiments on a variety of linear and nonlinear classifiers, all available through the ScikitLearn API (SKlearn) \cite{sklearn_api}. We point the reader to the Table \ref{table: models} in the Appendix for relevant literature discussing the origin and mathematics of each. We aim to perform the following:
\begin{enumerate}
    \item Generate RFI under the static case using Shap and EQ \ref{eq: Absolute Normalized Shap}.
    
    \item Generate RFI under the dynamic case using algorithm \ref{algo : 1} and EQ \ref{eq: ANWA}.
    
    \item Identify the similarity between (1) and (2) using EQs \ref{eq: cos} and \ref{fig:main_jac}.
    
    \item Offer analyses around noted differences in explanations under contrasting model specifications and classification metrics, differing feature sets/types, and elasticity of similarity as a function of dataset size.
\end{enumerate}

\subsection{Datasets}

We note several popular datasets used in Table \ref{table : datasets}. Most datasets are available via SKlearn, while the others can be sourced directly from the \hyperlink{https://archive.ics.uci.edu/ml/datasets}{UCI Machine Learning Repository} \cite{Dua:2019}, and were chosen due to their widespread use as benchmarking datasets in ML and DL to evaluate efficacy on a variety of performance metrics, as well as the variability in independent variables ranging from continuous only to mixtures of ordinal, nominal and continuous. We also introduce a synthetic fraud dataset built on real world transaction fraud data from J.P. Morgan to understand how our approach works on real-world data. 

\subsection{Performance Metrics}
\textit{Our} measure to compute feature importance under dynamic scenarios requires an arbitrary performance metric $m$ that measures $\hat Y^{test}$ against $Y^{test}$ before computing the ANWA via EQ \ref{eq: ANWA}. Accuracy is an oft-used metric in classification tasks but falters when dealing with datasets containing class imbalance (non-uniform distribution of the target variable), which resembles most real-world problems. We use Accuracy, Precision, Recall and F1-Measure throughout our analyses, defining and formalizing these in Table \ref{table: metrics}.
\begin{table}[h!]
\small
    \centering
    \begin{tabular}{P{1.5cm}P{2.25cm}P{2.25cm}}
    \hline
        \textbf{Metric} & \textbf{Definition} & \textbf{Notation}  \\ \hline
        
        Accuracy & Prop. of true preds amongst all preds  & $\alpha = \frac{TP+TN}{T P+T N+F P+F N}$ \\ \hline
        
        Precision & Prop. of true positive preds amongst all positive preds  & $p = \frac{TP}{TP + FP}$  \\ \hline
        
        Recall & Prop. of true positive preds amongst all actual positive instances & $r = \frac{T P}{TP + FN}$  \\ \hline
        
        F1 Measure & Harmonic Mean of Precision and Recall & $f1 = 2 \cdot \frac{p \cdot r}{p + r}$  \\ \hline
    \end{tabular}
    \caption{Commonly used metrics when dealing with a classification task $f:x \xrightarrow{} y $, where $y$ is a discrete target variable pertaining to a particular class. In the cases where the number of target variables exceeds 2, a weighted metric is required. The Iris and the Wine: datasets noted above require weighted, macro or micro averages for Precision, Recall and F1 measure calculations.}
    \label{table: metrics}
\end{table}

\subsection{Shap vs ANWA - A Drill Down}\label{sec: 4.3}
\begin{figure}[h!]
\centering
\includegraphics[scale=0.53]{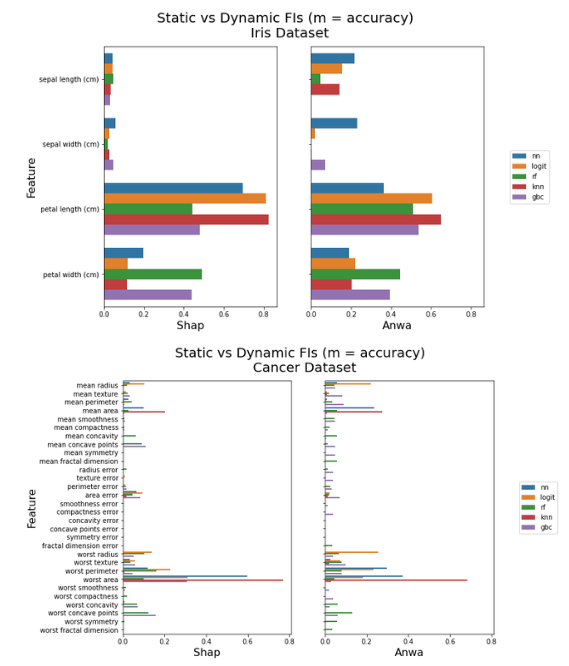}
\caption{Analysis conducted on the Iris \textbf{(TOP)} and Cancer \textbf{(Bottom)} datasets \cite{Dua:2019}.\textbf{ Left:} Absolute Normalized Shap Values for each feature. \textbf{Right:} Absolute Normalized Weighted Averages for each feature following our methodology for dynamic perturbation. We fix the arbitrary metric per EQ \ref{eq: ANWA} as $m =$ accuracy for simplification. The legend denotes abbreviations that follow the mapping: \textbf{'nn'} : Multilayer Perceptron Neural Network, \textbf{'svm'} : Support Vector Machine Classifier, \textbf{'logit'} : Logistic Regression, \textbf{'rf'} : Random Forest Classifier, \textbf{'knn'} : K-Nearest Neighbors Classifier, \textbf{'gbc'} : Gradient Boosted Trees Classifier.}
\label{fig:shapddd}
\end{figure}

Due to page limitations, we omit a complete, detailed drill-down on the measured RFIs for Shap against our methodology. We do, however, include a sample comparison displaying RFIs for a set of models on the Iris and Cancer datasets. 

What we see from these feature importance graphs (Iris), measuring the static (Shap) and dynamic (perturbation-based) scenarios, is the following:

\begin{itemize}
    \item Shap deems the feature 'sepal length' to be negligible in its effect on the models' ability to partition the classes. Most of the predictive power comes from 'petal length' and 'petal width'.

    \item Using our algorithm \ref{algo : 1} to perturb test set instances and generate RFIs, we show that 'sepal length' has a more recognizable influence on the model's outputs, which can be construed as contrasting explanations between the two methods. This is especially noticeable as we see the respective values drastically shift for the shallow neural network, where under the static case the global explanations emphasized petal length as being important, and under testing conditions, petal length was shown to have \textit{no} effect at all on the sensitivity of the model ('sepal length' took most of this difference in contributions).
\end{itemize}

We notice this as a recurring theme across our analyses of other datasets, where the explanations shift under the two scenarios, which raises questions about the trustworthiness of the model artifacts, particularly within highly regulated industries. If the models are not interpretable or harmonized in their interpretations, there is the potential for unintentional bias. We also include feature importance graphs for the Cancer dataset in Figure \ref{fig:shapddd}, where it becomes easier to notice how challenging a task creating 'reference narratives' can be in high dimensional spaces.

\subsection{Systematic Comparison}
While model artifact drill-downs may be necessary in some cases, our goal is to systematically quantify the similarity that we highlighted in \ref{sec: 4.3}. We dissect these results in Table \ref{table: mainresults}, where we show the similarity induced using various classification metrics on our algorithm across a suite of predictive models. 

In general, we notice that the choice of performance metric $m$ that we use to compute the feature importance via Eq \ref{eq: ANWA} has a marginal impact on the outcome, except in a few cases. Across all datasets, the average similarity across metrics ranges from [0.79, 0.84] with a standard deviation between [0.15, 0.22]. This variation reduces drastically if we remove the results from the Census dataset ($\mu = 0.88$, $\sigma = 0.13$). 

\begin{figure*}[hbt!]
\centering
\includegraphics[scale=0.65]{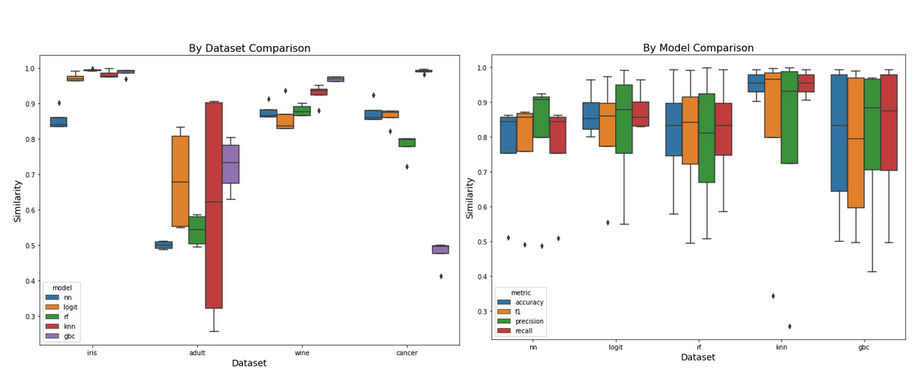}
\caption{\textbf{Left:} Boxplot displaying cosine similarity between Shap and ANWA by dataset. The colors map to specific predictive modeling techniques. \textbf{Right:} Boxplot displaying similarity between Shap and ANWA across all models, with the color mapping to the specific performance metric $m$ used.}
\label{fig:main}
\end{figure*}

The three datasets where the explanations align well under both static and dynamic forecasting are low dimensional, and contain only continuous features. The Census dataset is high dimensional and has a large number of boolean response variables. In that case, the RFIs for the continuous features are less emphasized and more diverse, i.e., when there are more explanatory variables, there is a lesser amount of unanimous consensus. This would be alarming in a deployed setting, like fraud detection, where the feature space is large and the feature types vary. This is not specific to Deep Learning algorithms, as it appears in similarity inconsistency across all four of the nonlinear estimators. The logistic regression model displays the highest average similarity, as well as the lowest variance. These results are further visualized in Figure \ref{fig:main}, and speak to a consensus between the two scenarios.
\begin{table*}[h!]
\small
\setlength\tabcolsep{4.5pt}
\makebox[\textwidth][c]{%
\begin{tabular}{|c|cccc|cccc|cccc|cccc|cccc|}
    \hline
    \multirow{2}{*}{} & \multicolumn{4}{c}{\textbf{NN}}  & \multicolumn{4}{c}{\textbf{Logit}} & \multicolumn{4}{c}{\textbf{RF}} & \multicolumn{4}{c}{\textbf{KNN}} & \multicolumn{4}{c|}{\textbf{GBC}} \\
    
    & $\alpha$ & $p$ & $r$ & $f$ & $\alpha$ & $p$ & $r$ & $f$ & $\alpha$ & $p$ & $r$ & $f$ & $\alpha$ & $p$ & $r$ & $f$ & $\alpha$ & $p$ & $r$ & $f$  \\
    
    \hline
         Adult  &  .51  &  .49  &  .51  &  .49  &  .80  &  .55  &  .83  &  .55  &  .58  &  .51  &  .59  &  .49  &  .90  &  .26  &  .90  &  .34  &  .69  &  .80  &  .77  &  .63  \\ \hline
         Cancer  &  .85  &  .92  &  .85  &  .86  &  .88  &  .82  &  .88  &  .87  &  .80  &  .72  &  .80  &  .80  &  .99  &  .98  &  .99  &  1.00  &  .50  &  .41  &  .50  &  .50  \\ \hline
         Iris  &  .83  &  .90  &  .83  &  .85  &  .96  &  .99  &  .96  &  .97  &  .99  &  1.00  &  .99  &  .99  &  .97  &  1.00  &  .97  &  .98  &  .99  &  .97  &  .99  &  .99  \\ \hline
         Wine  &  .86  &  .91  &  .86  &  .87  &  .83  &  .94  &  .83  &  .85  &  .86  &  .90  &  .86  &  .89  &  .94  &  .88  &  .94  &  .95  &  .97  &  .96  &  .97  &  .96  \\ \hline
    \hline
\end{tabular}
}
\caption{For each of the models we experiment with , we generate a cosine similarity measure (EQ \ref{eq: cos}) between the RFIs generated via Shap and by ANWA. This table shows the results when using accuracy, precision, recall or F1 measure as an input $m$ into Algorithm \ref{algo : 1}. $sim \xrightarrow{} 1$ means the explanations are similar, and as $sim \xrightarrow{} 0$ there exists disharmony between the static and dynamic cases.}
\label{table: mainresults}
\end{table*}

\subsection{Sample Size Analysis}
We want to see how well explanations align when models are allotted varying input sample sizes. For this, we isolate the Census dataset, as it has the largest total number of samples, and we fix $m = accuracy$ to be our metric of choice for computing the RFIs under a dynamic scenario. We set the size of the full dataset to be $|X| \in \{1000,2000,4000,8000,16000,32000\}$ and follow the same cleaning, splitting, and perturbation protocols as the previous step. 

\begin{table}[hbt!]
    \small
    \centering
    \setlength\tabcolsep{5pt}
    \begin{tabular}{|l|l|l|l|l|l|l|}
    \hline
        \textbf{$|X|$} & 1k & 2k & 4k & 8k & 16k & 32k \\ \hline
        \textbf{Similarity} & 0.70 & 0.56 & 0.49 & 0.74 & 0.86 & 0.87 \\ \hline
    \end{tabular}
    
    \caption{Subsample analysis on UCI Census Data using a Neural Network and $m = accuracy$.}
    \label{tab: mainsub}
    \end{table}

Table \ref{tab: mainsub} shows results from this experiment. We notice a  trend towards unified explanationsas the size of the sample pool grows, but variation in the similarities is too profound to assume a purely linear relationship ($R^2 = 0.58$). While increasing the size of the dataset has a corresponding effect on the model's performance, we do not see enough evidence to support that it implies more similar explanations. More work is required to understand the relationship between the two, as well as the performance-interpretability trade-off that comes when imposing likeness as a constraint on the training process.

\subsection{Ranked Similarity}
Following from EQ \ref{metric: jacc}, we aim to quantify how often the top-k features resultant from the two explanation methods align. In highly regulated areas, like finance, we may not be as concerned with uniform harmony under dynamic forecasting scenarios, but rather with alignment amongst top features that the model identifies as being important.

Using Shap as our baseline, we compute the rank of each feature in the feature set by absolute normalized Shap value, and compare them to the rankings pursuant from our method, ANWA, with varying $k \in K$, where k is the size of the slice over the sorted set. 

\begin{figure}[hbt!]
\centering
\includegraphics[scale=0.48]{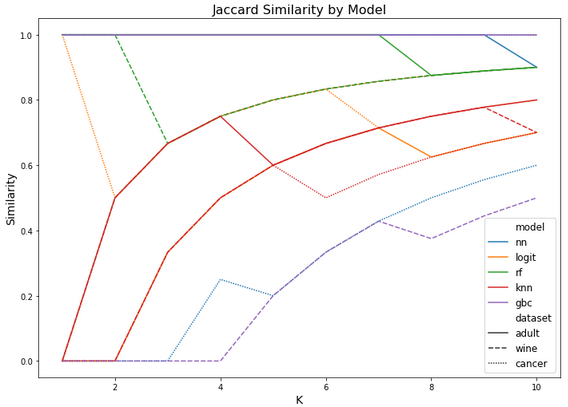}
\caption{Jaccard Similarity by Model, by Dataset. The similarity is measured by the intersection of two sets against their union. The two sets contain the top-k features for each method under static and dynamic XAI.}
\label{fig:main_jac}
\end{figure}

Iris only contains four features, so we exclude that dataset from this analysis. We see a general monotonic trend across all estimators, for all datasets. At the $k =10$ step, all models have a Jaccard Similarity greater than 0.5 ($\mu = 0.84$, $\sigma = 0.16$). While a large $k$ can artificially inflate the similarity measure as $k \xrightarrow{} |m|$, we find the general trend to be true for the datasets with a high feature dimension (Breast Cancer and UCI Census). Results are visualized in Fig \ref{fig:main_jac}, and pseudocode for the heuristic we use to generate this measure is included in algorithm \ref{algo : 2} in the Appendix.

\subsection{Explainability in the Wild}
We turn our attention to a synthetic dataset provided by J.P. Morgan \cite{jpm1, jpm2}, with the underlying scope of the task being to classify fraudulent transactions from genuine ones (all data cleaning notes can be found in the Appendix). The testing suite of models outperform a majority vote baseline (50\%) by $\sim 20\%$ without hyperparameter tuning, whereas the full dataset had a baseline of $98\%$ (i.e., $98\%$ of the samples belong to the non-fraudulent class). The Cosine Similarity between the two explanation vectors, along with the model accuracy on the testing set, are shown in Table \ref{table: rwd}.

\begin{table}[h!]
\small
    \centering
    \begin{tabular}{|l|l|l|}
    \hline
        \textbf{Model} & \textbf{Accuracy} & \textbf{Cos Similarity} \\ \hline
        Neural Network & 0.58 & 0.43 \\ \hline
        Logistic Regression & 0.6 & 0.9 \\ \hline
        Random Forest & 0.65 & 0.88 \\ \hline
        K-Nearest Neighbors & 0.58 & 0.99 \\ \hline
        Gradient Boosting & 0.66 & 0.97 \\ \hline
    \end{tabular}
    \caption{Model Accuracy and Cosine Similarity between explanation techniques on the J.P. Morgan synthetic dataset.}

    \label{table: rwd}
\end{table}

While the models are not fully tuned, there are notable similarities to the SRGs reported from Shap against our method, which does not require the full retraining of models on the $S \subseteq F$ possible feature subsets. The lone outlier is in the case of the Deep Neural Network, which is under-powered by a limited amount of positive (fraudulent) examples in the dataset. Using JS as a proxy for the harmony of explanations, we note that a lower Cosine Similarity between explanation vectors does not necessarily imply low congruence. For $k \in \{1, \dots, 10\}$ we see the mean JS, $\frac{1}{|k|}\sum_i^{|k|}J(k)$, near 80\% ($\sigma = 0.14$) with $k \leq \frac{1}{4}|X(j)|$. Disharmony between the explanation vectors can be subjective based on the intended use-case of understanding the model artifacts. While using JS may allow us to understand the most important features, a model that exploits $\frac{1}{|X(j)|}$ of the feature space for more than half of its predictive power (as was the case for Transaction\_Amount to detect fraud here) should be monitored, or further tuned.

We believe that these experiments work to further validate the explanation models resultant from the Shap attribution generation process by showing a consensus between methods. Further, we believe this to be a novel framework for dissecting general predictive models and their ensuing explanations, with a lesser computational burden than experienced with generating SRGs.

\section{Conclusion}
XAI offers a paradigm shift towards interpretability and explainability required in many fields utilizing ML and AI. We have introduced a taxonomy classifying two unique instances of XAI, 'dynamic' and 'static' cases \cite{dolk2020model}, formulated harmony as a measure of the distance between explanations of these cases (Cosine and Jaccard Similarity), and employed a  perturbation-based algorithm \ref{algo : 1} to systematically quantify it on several models, metrics, and datasets.  

We show a moderate to high level of consensus among methodological views, one looking towards attribution values generated from the training data, and one which looks towards the testing data to validate it on \emph{potentially} out of sample datapoints and distributions. Namely, for low dimensional datasets, we see that perturbations, even extreme ones, do not cause drastic shifts in the magnitude of feature importance's, or their general orderings. This proposed framework \textit{begins} to shine a light on questions sparsely asked in XAI, such as "How well does the explanation model hold up in production?", and "Can I trust this explanation model in a deployed setting, without unintentionally amplifying bias and unfairness?", by showing similarity between the explanations generated in the static case, and in the dynamic case, where the distributions for features could shift, or drift, rapidly. Our proposed method is also flexible, intuitive, and easy to implement. 

We believe that this work can be catalyzed to answering those questions, and towards facilitating the generation of reference narratives as a way to answer the question “what’s going on here?” in any particular model and decision-making setting. Also, we believe this work to be essential to the overall model life cycle, where checks-and-balances are needed to show shortcomings of over-exposed bias from the model artifacts.  

\subsection{Future Work}
XAI is very much an open research problem, stemming from widespread industrial adoption of sophisticated algorithms, and the importance for them to pass tests concerning bias, fairness, and transparency. In the future, we intend to explore the framework that we have specified here for datasets under conditions faced by highly regulated industries, like finance, where mitigating bias in explanations is of the utmost importance. We believe the ideas expressed here could be relevant to non-tabular, unstructured data, where it could be a useful addition to the growing literature on XAI in its quest to dispel explanation uncertainty in tasks like Natural Language Processing, Computer Vision, and Graph Machine Learning. Where there is a machine learning model, there is a need for viable explanation techniques that practitioners can rely on to illuminate the black-box nature of a class of highly parameterized, nonlinear functions.


\bibliographystyle{ieeetr}
\bibliography{sample}

\appendix
\section{Appendix}
\section{Appendix}\label{app: appendix}

\subsection{Algorithm}
We note pseudocode for our dynamic perturbation algorithm here and include code written in python linked to our repository. Additionally, we note pseudocode for our means of comparing static and dynamic RFIs using Jaccard Similarity.

\begin{algorithm}[h!]
\small
\caption{Perturbation($f, m, X,Y, j, P$)}    
        $W$ = weighted average\\
        $m$ = performance metric \\
        $X \in R^{n x m}$ \\
        $Y \in R^n$ \\
        $p_i \in P$ \\
        $j \in |m|$ \\
        $(X^{train}, X^{test}, Y^{train}, Y^{test}) \in X,Y$\\
        $X^{Cont}, X^{Cat} \in X^{test}$ \\
    
    $W = 0$\\
    $f \xleftarrow{} f(X^{train}, Y^{train})$ \\ 
    \For{$p_i \in P$}  {
    //Outer loop \\
    \For{$j \in |m|$} {
        //Inner loop\\
        //Perturb $jth$ feature by $p_i$\\
        if $x \in X^{Cont}$: \\
        \quad $X^{test'}(j,p_i) = X^{test}(j) * p_i$\\
        if $x \in X^{Cat}$: \\
        \quad if $p_i \geq 1$:\\
        //Activate inactive observations for feature $x$\\
        \quad \quad $X^{test'}(j,p_i) = X^{test}(j = 1) * (2-p_i)$\\
        //Deactivate active observations for feature $x$\\
        \quad else:  \\
        \quad \quad $X^{test'}(j,p_i) = X^{test}(j = 1) * (1-p_i)$\\
        //apply trained model f\\
        $\hat y = f(X^{test'})$\\
        //generate performance of model\\
        $w_i = (1- |1-p_i|)$ \\
        $W(j) += \frac{w_i \cdot m(\hat y, y)}{\sum_i^P w_i }$\\
    }
    \textbf{End for}
    }
    \textbf{End for}\\
    \textbf{Return}: Weighted average $W$ for perturbing feature $j$ for each $p_i \in P$

\label{algo : 1}
\end{algorithm}

\begin{algorithm}[h!]
\small
\caption{Jaccard($dataset, model, k$)}    
        $k$ =slice range\\

    svs =  generate\_shap\_values($\cdot$)\\
    anwa = generate\_anwa\_values($\cdot$)\\
    
    top\_k\_shap = rank(sort(svs))[:k]\\
    top\_k\_anwa = rank(sort(anwa))[:k]\\

    intersection = $top\_k\_shap \cap top\_k\_anwa$\\
    union = $top\_k\_shap \cup top\_k\_anwa$\\
    
    Jaccard = $\frac{intersection}{union}$\\
    \textbf{Return} Jaccard

\label{algo : 2}
\end{algorithm}

\subsection{Models}
We include some cursory information on each of the predictive models used in our experiments in Table \ref{table: models}. Models were \textbf{not} trained with extensive hyperparameter tuning aimed at driving marginal performance improvements. SVM was excluded from our analysis as Shap value generation was a bottleneck on the hyperplane-deriving algorithm. Specific configurations can be found in our source code (available later).
\begin{table}
    \small
    \centering
    \begin{tabular}{|l|l|l|}
    \hline
        Model Name & Abbrev. & Refs\\ \hline
        Deep Neural Network & nn & \cite{Rosenblatt1963PRINCIPLESON, 2091}  \\ \hline
        Support Vector Machine & svm & \cite{noble2006support,wang2005support} \\ \hline
        Logistic Regression & logit & \cite{peng2002introduction, wright1995logistic} \\ \hline
        K-Nearest Neighbors & knn & \cite{peterson2009k} \\ \hline
        Random Forest & rf & \cite{liaw2002classification}  \\ \hline
        Gradient Boosted Trees & gbc & \cite{gbc}  \\ \hline
    \end{tabular}
    \caption{Abbreviation to Model map used in this paper. References to seminal work, or surveys, on each can be found here. All models were trained, validated and tested using the Sklearn API, and all data cleaning, splitting, and modification followed suit.}
    \label{table: models}
\end{table}

\subsection{Datasets}
\label{datasets}
In general, there was little in the way of data cleaning. The J.P. Morgan Fraud dataset, being a real-world, synthetic set, was downsampled to have a uniform class distribution to deal with imbalance, and certain nominal features with little predictive power were omitted to limit exploding dimensionality. The transaction timestamp feature was parsed and hour, day of week, and month were used as categorical features. Iris, Wine, and Cancer are all continuous datasets. UCI Census contains categorical variables, and those were preprocessed using pandas dataframe operations. Each categorical variable was transformed into a one-hot boolean response vector. All datasets were split into train and test sets using the Sklearn API, and generally using an 80-20 split, except for specific analysis. These datasets were randomly shuffled before splitting.

\subsection{Shap}
Our work is dependent on Shap \cite{Lundberg_Lee_2017} to derive baseline RFIs. RFIs in the static case can be dependent on parameter configurations. As computing Shap values over a full dataset with high dimensionality (in the case of binary expansion) can be expensive, we summarize the training/reference dataset into informative samples using the shap.kmeans call.

\end{document}